\documentclass[12pt]{article}

\usepackage{sbc-template}

\usepackage{graphicx,url,booktabs,bm,amsmath,xcolor,array}
\newsavebox\CBox
\def\textBF#1{\sbox\CBox{#1}\resizebox{\wd\CBox}{\ht\CBox}{\textbf{#1}}}

\usepackage[utf8]{inputenc}  

\title{Scaling Up ESM2 Architectures for Long Protein Sequences Analysis: Long and Quantized Approaches\footnote{This paper was presented at BSB 2024. DOI: \url{https://doi.org/10.5753/bsb.2024.244804}}}

\author{Gabriel Bianchin de Oliveira\inst{1}, Helio Pedrini\inst{1}, and Zanoni Dias\inst{1}}

\address{Institute of Computing, University of Campinas, Campinas, SP, Brazil
  \email{\{gabriel.oliveira, helio, zanoni\}@ic.unicamp.br}
}

\begin{document} 

\maketitle

\begin{abstract}
Various approaches utilizing Transformer architectures have achieved state-of-the-art results in Natural Language Processing (NLP). Based on this success, numerous architectures have been proposed for other types of data, such as in biology, particularly for protein sequences. Notably among these are the ESM2 architectures, pre-trained on billions of proteins, which form the basis of various state-of-the-art approaches in the field. However, the ESM2 architectures have a limitation regarding input size, restricting it to 1,022 amino acids, which necessitates the use of preprocessing techniques to handle sequences longer than this limit. In this paper, we present the long and quantized versions of the ESM2 architectures, doubling the input size limit to 2,048 amino acids.
\end{abstract}

\section{Introduction}
Transformer-based models have become state-of-the-art in various Natural Language Processing (NLP) tasks, such as context analysis, text generation, and translation. Recently, tools that utilize Transformers, such as ChatGPT\footnote{\url{https://chatgpt.com}} and Copilot\footnote{\url{https://copilot.microsoft.com}}, have become instrumental in assisting users with their tasks.

The success of architectures that employ Transformers stems from attention modules, capable of learning the relationships between words in a sentence autonomously (self-attention)~\cite{vaswani2017attention}. In the pre-training phase, these models are exposed to the context of the language they are being trained in, which typically consists of collections ranging from millions to billions of documents, enabling them to learn and adapt to the nuances of the language. After this initial stage, which takes a substantial amount of time and requires significant processing power, users can fine-tune the model for specific tasks.

Following the success in Natural Language Processing, this approach is also being applied in other contexts, such as images~\cite{arnab2021vivit,dosovitskiy2020image} and audio~\cite{ao2021speecht5}. In the biological domain, several Transformer-based architectures have also been developed, becoming state-of-the-art in tasks such as protein structure prediction~\cite{abramson2024accurate,lin2023evolutionary}, protein representation extraction~\cite{elnaggar2021prottrans}, biological article analysis~\cite{lee2020biobert}, and DNA sequence analysis~\cite{zhou2023dnabert}.

Considering the approaches for proteins, ESM family architectures, developed by the MetaAI group, with the most recent version being ESM2~\cite{lin2023evolutionary}, are among the state-of-the-art approaches in various tasks, such as protein function prediction~\cite{zhapa2024predicting}, protein family annotations~\cite{vitale2024evaluating}, and protein sequence conservation~\cite{yeung2023alignment}. The original ESM2 architecture has restrictions regarding the maximum sequence size of 1,022 amino acids, which, together with the~\texttt{CLS} token, used to indicate the beginning of the sequence and utilized in classification tasks, and the~\texttt{EOS} token, used to indicate the end of the sequence, total a 1,024-token input limit. However, there are protein sequences larger than this maximum size, forcing the approaches to use techniques such as truncation up to this limit, excluding larger proteins, or treatments with sliding window techniques to deal with longer sequences.

In this paper, we introduce the long versions of ESM2 architectures, which can process proteins with up to 2,048 amino acids without the need for additional preprocessing. Besides the standard long versions, we also present the quantized long versions, referred to simply as quantized, which apply quantization to reduce memory space required for model loading and to accelerate inference time.

Quantization is a technique commonly used in neural networks, including Transformers, to decrease the model's precision from 32-bit floating point to lower bit-width representations, such as 8-bit and 4-bit integers. This process significantly reduces memory usage and computational requirements, often with minimal impact on model accuracy. Consequently, model loading times and inference speeds are improved, making it a desirable option for deploying large models in resource-constrained environments.

During our evaluation, we assessed the ESM2 long and ESM2 quantized architectures for the task of protein function prediction. In most cases, these architectures demonstrated superior performance compared to the standard ESM2 architecture.

The remainder of the paper is organized as follows. In Section~\ref{sec:methodology}, we describe the proposed architectural adaptation of ESM2 models to deal with sequences up to 2,048 amino acids. In Section~\ref{sec:results}, we evaluate and discuss the results for the protein function prediction task using the embeddings extracted from the long and quantized architectures and compare them with the standard ones. In Section~\ref{sec:conclusions}, we present the main aspects of our work and indicate possible points for future research.

\section{Methodology}
\label{sec:methodology}
\begin{figure}[!htb]
\centering
\begin{minipage}{0.5\textwidth}
\centering
\includegraphics[width=\textwidth]{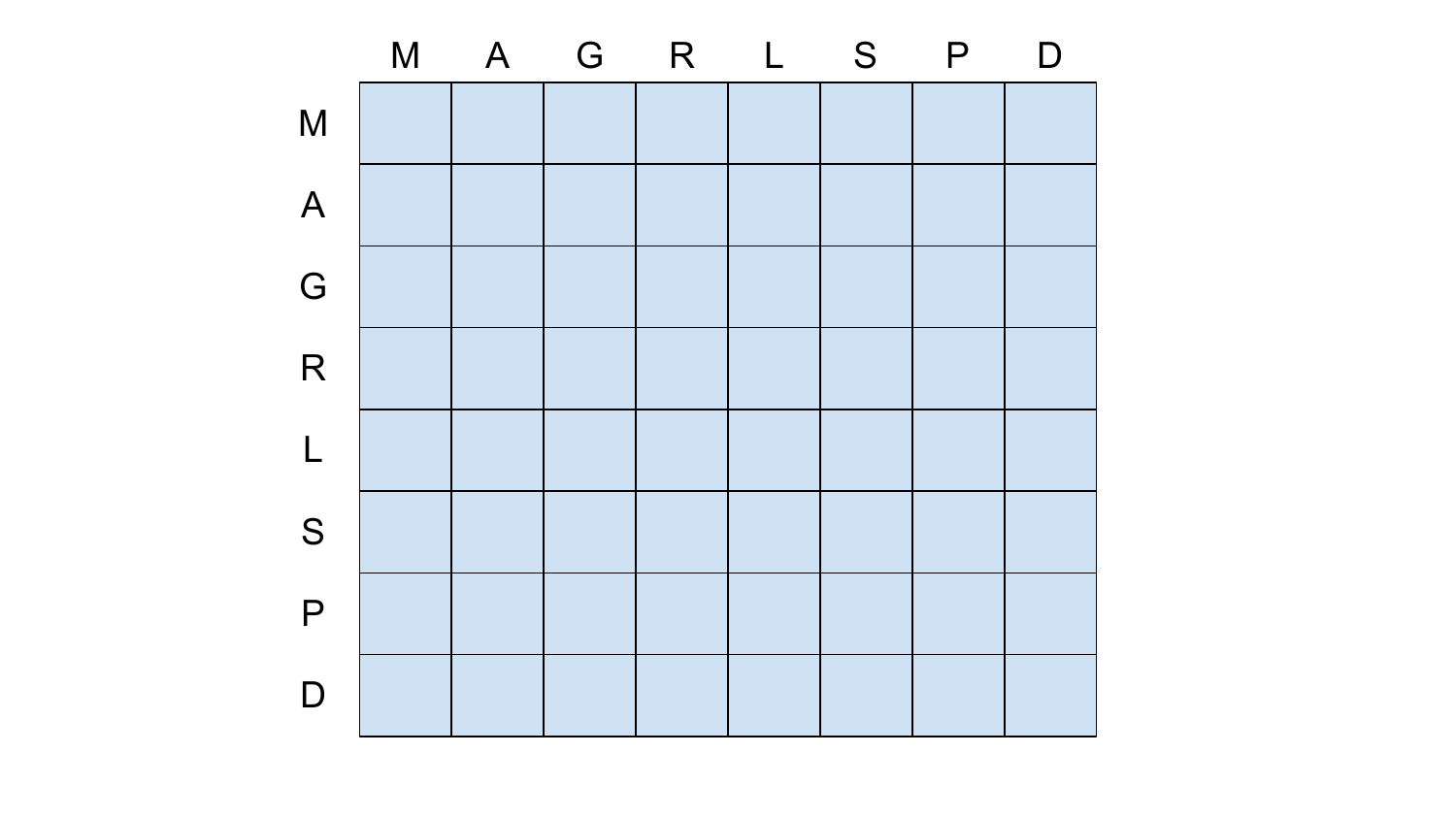}
\par (a) Global.
\end{minipage}%
\begin{minipage}{0.5\textwidth}
\centering
\includegraphics[width=\textwidth]{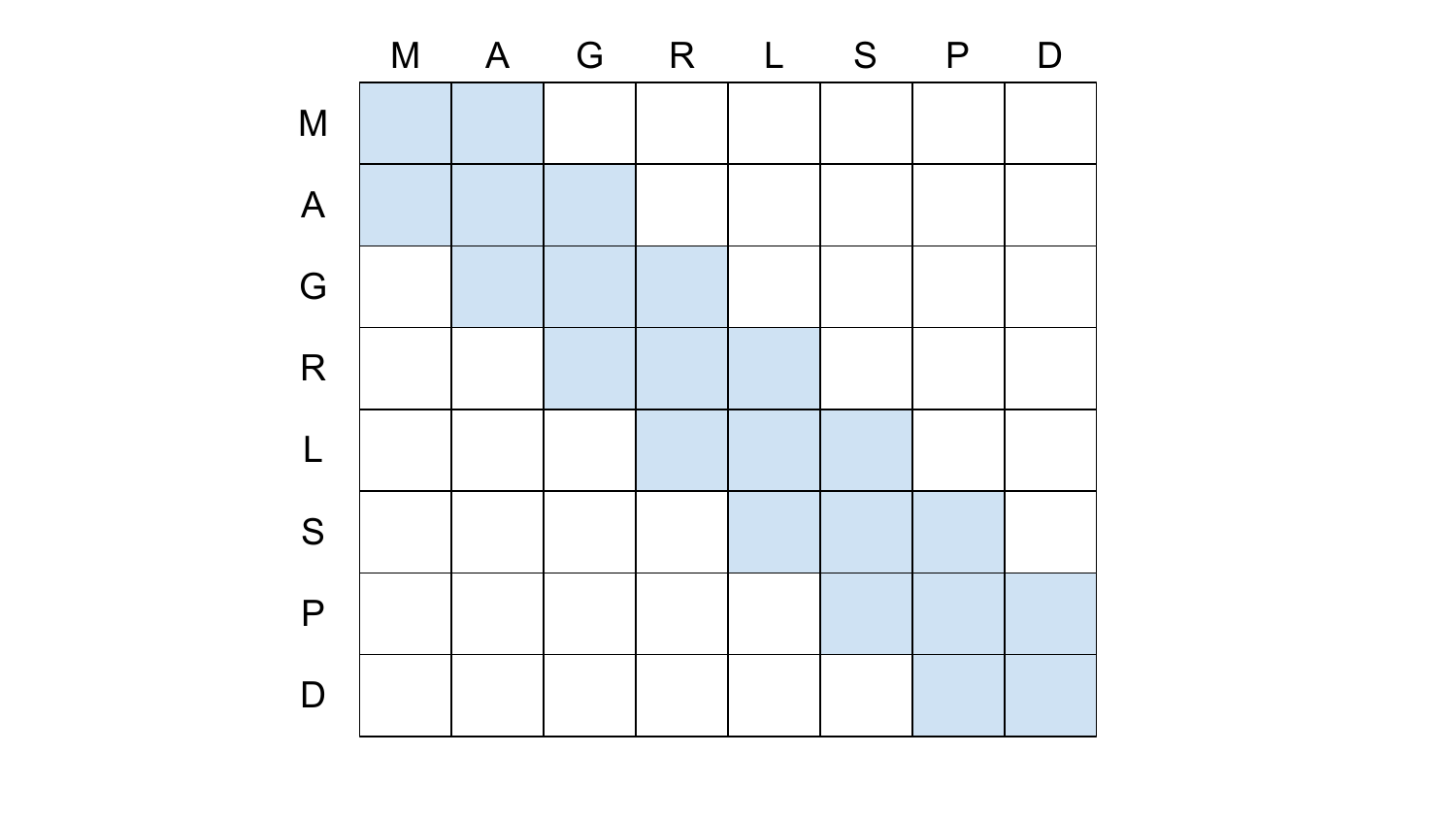}
\par (b) Local.
\end{minipage}
\caption{Self-attention mechanisms. In the global self-attention mechanism, each amino acid examines all the amino acids in the sequence. In the local self-attention mechanism, each amino acid examines the amino acids within a specific window.}
\label{fig:attention}
\end{figure}

ESM2 architectures have configurations highlighted in Table~\ref{tab:config}. Each of these architectures employs the concept of self-attention memory modules with global mechanism, that is, each token (amino acid,~\texttt{CLS}, or~\texttt{EOS}) examines all other tokens in the sequence, as depicted in Figure~\ref{fig:attention}. Taking into account memory and computational processing, the ESM2 architectures perform attention calculation in $\mathcal{O}(n^2)$, where $n$ is the sequence length.

\begin{table}[!t]
\centering
\caption{Configuration of ESM2 architectures. Each architecture in the ESM2 family has $n$ stacked layers, ranging from 6 in T6 up to 48 in T48.}
\begin{tabular}{lrrrrrr}
\toprule
& \multicolumn{1}{r}{\textbf{T6}}  & \multicolumn{1}{r}{\textbf{T12}} & \multicolumn{1}{r}{\textbf{T30}} & \multicolumn{1}{r}{\textbf{T33}} & \multicolumn{1}{r}{\textbf{T36}} & \multicolumn{1}{r}{\textbf{T48}}\\
\midrule
Number of Layers & 6 & 12 & 30 & 33 & 36 & 48 \\
Attention Heads & 20 & 20 & 20 & 20 & 40 & 40 \\
Embedding Dimension & 320 & 480 & 640 & 1,280 & 2,560 & 5,120 \\
\bottomrule
\end{tabular}
\label{tab:config}
\end{table}

Inspired by LongFormer~\cite{beltagy2020longformer}, we modified the attention mechanisms of the ESM2 architectures to local form, in which each token considers only the other tokens within a window of size $k$, as depicted in Figure~\ref{fig:attention}. Consequently, the computational and memory complexity takes the form $\mathcal{O}(nk)$, where $n$ is the sequence length and $k$ is the window size. 

To implement this adaptation, we copied the context representation of the ESM2 architectures to 2,050 positions, with 2,048 allocated for the amino acids and 2 positions for special tokens (\texttt{CLS} and~\texttt{EOS}). We adopted this approach based on the results from Beltagy~\textit{et al.}~\cite{beltagy2020longformer}, which demonstrated that context copying is more effective than random initialization. In addition to altering the context representation, we also modified the attention modules from global to local mechanisms.

Concerning the window size of attention, we maintained the window size at 1,024 and increased the sequence limit to 2,048 amino acids. Consequently, even though this increases the memory requirement compared to the original models, which only had an input size of 1,024 tokens, the memory needed for adapting the model to accommodate an input size of 2,050 tokens (up to 2,048 amino acids,~\texttt{CLS}, and~\texttt{EOS}) with global attention analysis was halved using our approach.

In addition to the long version, we transformed ESM2 long architectures into quantized versions. For this, we carried out the same process described for the long version, but during the architecture adaptation and pre-training stage, we loaded the models in the~\texttt{int4} format~\cite{dettmers2212case}, using LoRA~\cite{yu2023low} and~\texttt{bfloat16} computation type. Unlike the standard representation of machine learning models, which is~\texttt{float32}, representing each weight and network activation by 32 float values, the~\texttt{int4} version performs this representation with only 4 integer values, reducing the memory required to load models by up to 8 times, while at a cost in terms of model performance.

Following the modifications to the architecture for the long and quantized versions, we pre-trained the networks considering all proteins (569,793) available in July 2023 in the UniProt database, Swiss-Prot version~\cite{uniprot2023uniprot}. We opted for this version given that the proteins in this set have been reviewed by laboratory methods compared to UniProtKB-TrEMBL. During this stage, we trained the models for 5 epochs, with a learning rate of $10^{-5}$ and the AdamW~\cite{loshchilov2017decoupled} optimizer.

Table~\ref{tab:config-memory} presents the amount of memory required (in MB) to load each model. The memory requirement for small quantized models, such as T6, exceeds that of the standard and long configurations. However, as the model size increases, quantization proves to be memory-efficient, reducing the required memory by approximately four times for the largest ESM2 architecture that has both long and quantized versions (T33).

\begin{table}[!t]
\centering
\caption{Memory required to load each ESM2 architecture (in MB). Each architecture in the ESM2 family has $n$ stacked layers, ranging from 6 in T6 up to 48 in T48.}
\begin{tabular}{lrrrrr}
\toprule
& \textbf{T6}  & \textbf{T12} & \textbf{T30} & \textbf{T33} & \textbf{T36}\\
\midrule
Standard & 31 & 136 & 595 & 2,673 & 11,643  \\
Long & 40 & 171 & 746 & 3,338 & -  \\
Quantized & 314 & 328 & 384 & 664 & 1,750 \\
\bottomrule
\end{tabular}
\label{tab:config-memory}
\end{table}

Due to computational limitations, we were unable to transform the ESM2 T36 architecture into the long version, nor the ESM2 T48 architecture into both the long and quantized versions. All the ESM2 long and quantized architectures are available in our HuggingFace webspace\footnote{https://huggingface.co/gabrielbianchin}.

\section{Results and Discussion}
\label{sec:results}
In order to assess the embedding representations of both the long and quantized versions of ESM2 architectures, we conducted an evaluation with respect to the task of protein function prediction.

With the recent advancements of the past decades, such as next-generation sequencing, numerous proteins have had their amino acid sequences defined by laboratory methods. However, determining the functions that each of these proteins performs remains quite costly, considering both the time and the financial resources required for this type of laboratory analysis. As a result, various computational methods have been proposed to reduce the gap between proteins that have a defined sequence but lack an annotated function~\cite{cao2021tale,chua2024protgoat,kulmanov_deepgoplus:_2019,oliveira2023temprot,oliveira2024protein,zhu2022atgo}.

Protein functions are typically classified using Gene Ontology (GO)~\cite{ashburner_gene_2000}. This approach divides function annotation into three ontologies: Biological Process Ontology (BPO), which evaluates the overall process in which the protein is involved; Cellular Component Ontology (CCO), which indicates the location where the protein is executing its function; and Molecular Function Ontology (MFO), which analyzes the function performed at the molecular level. These three ontologies are organized in a directed acyclic graph, where deeper terms are more specific, while terms closer to the root term are more generic. Thus, if a protein performs a more specific function, it also performs all terms up to the root term, following the true path rule~\cite{valentini2010true}. Moreover, each protein can perform more than one function at the same time, even if these functions have no shared ancestral terms. Due to this nature, the problem of classifying protein functions is considered a multi-label classification by computational methods.

\begin{figure}
\centering
\includegraphics[width=\textwidth]{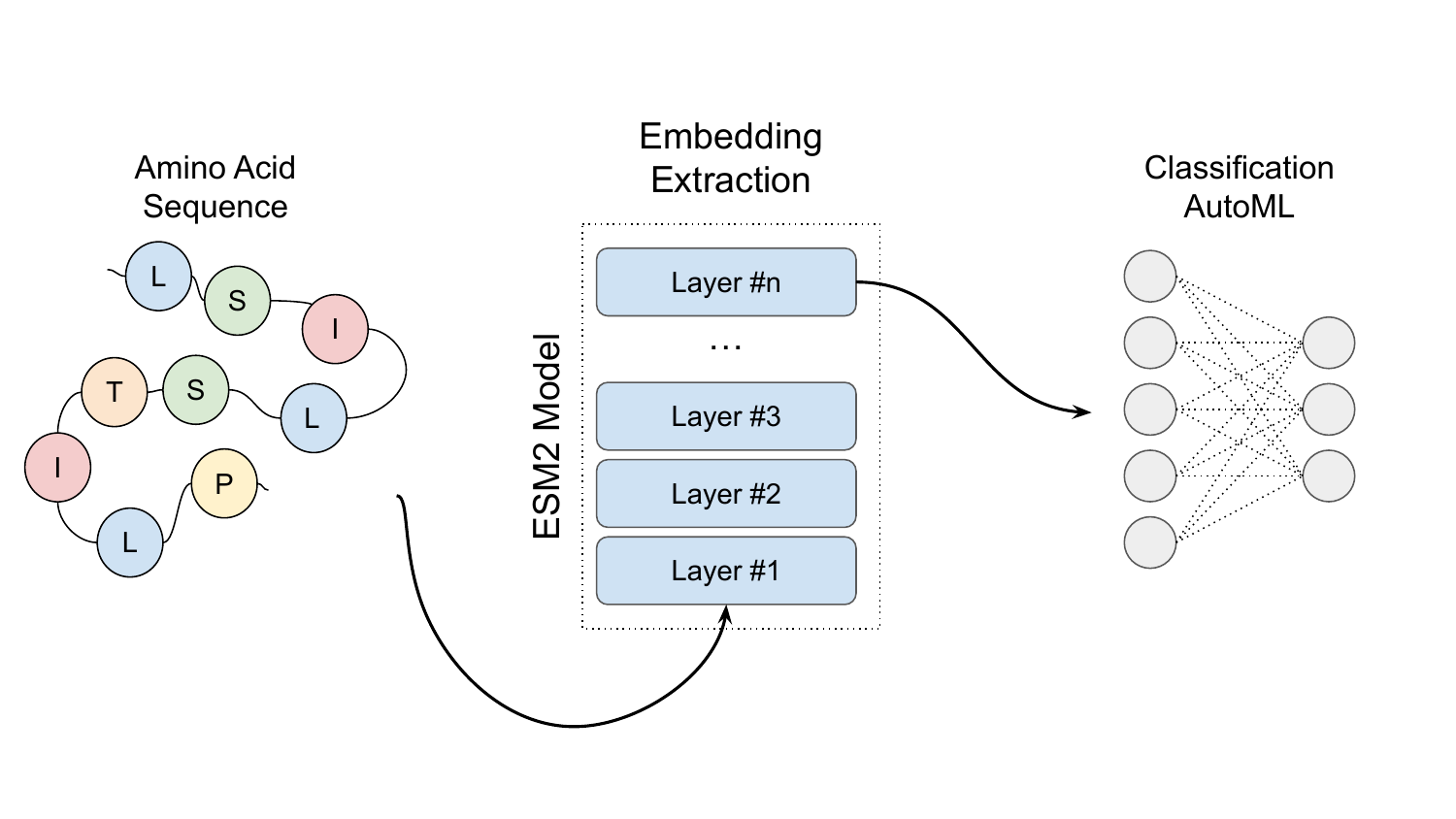}
\caption{Pipeline for evaluating protein embeddings from ESM2 architectures. The method receives the amino acid sequence as input. Then, the features from the last layer of the backbone are used to train a classifier. During the classification step, the best classification model is identified using AutoML.}
\label{fig:pipeline}
\end{figure}

To evaluate the representations (embeddings) extracted from ESM2 architectures, we employed the pipeline described by Oliveira et al.~\cite{oliveira2024protein}, illustrated in Figure~\ref{fig:pipeline}. In the initial stage, referred to as embedding extraction, we extracted embeddings from the last layer of each architecture for every protein in the training, validation, and test sets. Due to the length of the sequences and the maximum input size of the standard, long, and quantized architectures, if a protein sequence exceeds 1,022 (standard) or 2,046 (long and quantized) amino acids, we applied the sliding window technique to segment the sequence into non-overlapping slices that fit within the models' maximum input size. For example, if a protein has 3,000 amino acids, for the standard configuration, there will be two slices of 1,022 amino acids (the first from amino acid 1 to amino acid 1,022, and the second from amino acid 1,023 to amino acid 2,044) and one slice with 956 amino acids (from amino acid 2,045 to 3,000). For the long and quantized versions, there will be two slices, one with 2,046 amino acids (from amino acid 1 to 2,046) and another with 954 amino acids (the remaining ones).

After extracting the embeddings, if a protein was separated into slices, we aggregated all representations by averaging the feature vectors position-wise. If a protein sequence length was less than the model input limit, no preprocessing steps were applied. At the end of this process, each protein in the training, validation, and test sets is represented by real values, with the representation vector matching the embedding dimension specified in Table~\ref{tab:config}.

With the extracted embeddings, for each scenario -- considering architecture (T6, T12, T30, or T36), input size (standard, long, or quantized), and ontology (BPO, CCO, or MFO) -- we obtained a classifier using AutoML from the AutoKeras~\cite{JMLR:v24:20-1355} package with 50 trials, selecting the best classifier for each configuration based on the validation set. Finally, the best classifier identified for each scenario was evaluated on the test set.

The dataset utilized for the evaluation of ESM2 embeddings was chosen as outlined in the work of Oliveira et al.~\cite{oliveira2024protein}, which is derived from the CAFA5 challenge~\cite{cafa-5-protein-function-prediction}. The number of proteins in the training, validation, and test sets, as well as the number of terms in each ontology, are detailed in Table~\ref{tab:dataset}. With respect to proteins consisting of more than 1,024 amino acids, this dataset comprises approximately 12\% for BPO, 11\% for CCO, and 10\% for MFO. In the case of proteins with more than 2,048 amino acids, it is approximately 2\% for each ontology.

\begin{table}[!t]
\centering
\caption{Number of proteins and terms for BPO, CCO, and MFO.}
\begin{tabular}{lrrr}
\toprule
& \multicolumn{1}{r}{\textbf{BPO}}  & \multicolumn{1}{r}{\textbf{CCO}} & \multicolumn{1}{r}{\textbf{MFO}}\\
\midrule
Training & 73,768 & 74,328 & 62,909 \\
Validation & 9,221 & 9,292 & 7,864 \\
Test & 9,221 & 9,292 & 7,864 \\
\cline{1-4}
Terms & 500 & 498 & 499 \\
\bottomrule
\end{tabular}
\label{tab:dataset}
\end{table}

As an evaluation metric, we utilized $F_{\max}$, which is the most commonly employed metric in the task of protein function prediction~\cite{radivojac2013not,zhou_cafa_2019}. $F_{\max}$ assesses the maximum $F$-score considering the thresholds $\tau$ ranging from 0.01 to 1.00, applying the harmonic mean between precision and recall at each $\tau$. Equations~\ref{eq:f1max},~\ref{eq:precision}, and~\ref{eq:recall} present $F_{\max}$, precision at $\tau$ (denoted by $\operatorname{pr}(\tau)$), and recall at $\tau$ (denoted by $\operatorname{rc}(\tau)$). In these formulas, $T_i$ represents the ground-truth of a protein $i$, $P_i(\tau)$ is the set of terms predicted for a protein $i$ at a threshold $\tau$, $m(\tau)$ indicates the number of proteins with at least one term predicted with a score equal to or greater than $\tau$, and $n$ is the number of proteins in the evaluation set.

\begin{equation}
\label{eq:f1max}
 F_{\max} = \max_{\tau} \left\{ \frac{2 \times \operatorname{pr}(\tau) \times \operatorname{rc}(\tau)}{\operatorname{pr}(\tau) + \operatorname{rc}(\tau)} \right\}
\end{equation}
\begin{equation}
\label{eq:precision}
\operatorname{pr}(\tau) = \frac{1}{m(\tau)} \sum^{m(\tau)}_{i=1} \frac{\left | P_i(\tau) \cap T_i \right |}{\left | P_i(\tau) \right |}
\end{equation}
\begin{equation}
\label{eq:recall}
\operatorname{rc}(\tau) = \frac{1}{n} \sum^{n}_{i=1} \frac{\left | P_i(\tau) \cap T_i \right |}{\left | T_i \right |}
\end{equation}

The results of the models using ESM2 standard, long, and quantized embeddings for the protein function prediction task on the test set are presented in Table~\ref{table:results}. These results indicate that the optimal values, or the highest results for each type of architecture, from ESM2 T6 to ESM2 T33, are attained by long and/or quantized architectures, with the exception in ESM2 T30 for BPO. With respect to ESM2 T36, the quantized version achieved the best results for CCO, while the standard architecture surpassed it in the other two ontologies.

\begin{table}[!t]
\centering
\caption{$F_{\max}$ of ESM2 standard, long and quantized embeddings on the test set.}
\begin{tabular}{lp{1.5cm}p{1.5cm}p{1.5cm}}
\toprule
\textBF{Method} & \textBF{BPO} & \textBF{CCO} & \textBF{MFO} \\
\midrule
\textbf{ESM2 T6} & & & \\
\ \ \ Standard & 0.505 & 0.723 & 0.754 \\
\ \ \ Long & \textBF{0.509} & \textBF{0.733} & \textBF{0.757} \\
\ \ \ Quantized & 0.498 & 0.727 & 0.744\vspace{0.15cm} \\

\textbf{ESM2 T12} & & & \\
\ \ \ Standard & 0.505 & 0.728 & 0.762 \\
\ \ \ Long & \textBF{0.532} & \textBF{0.734} & \textBF{0.778} \\
\ \ \ Quantized & 0.505 & 0.729 & 0.777\vspace{0.15cm} \\

\textbf{ESM2 T30} & & & \\
\ \ \ Standard & \textBF{0.539} & 0.739 & 0.770 \\
\ \ \ Long & 0.527 & 0.742 & 0.766 \\
\ \ \ Quantized & 0.509 & \textBF{0.743} & \textBF{0.778}\vspace{0.15cm} \\

\textbf{ESM2 T33} & & & \\
\ \ \ Standard & 0.540 & 0.736 & 0.773 \\
\ \ \ Long & 0.512 & \textBF{0.751} & 0.782 \\
\ \ \ Quantized & \textBF{0.549} & 0.747 & \textBF{0.783}\vspace{0.15cm} \\

\textbf{ESM2 T36} & & & \\
\ \ \ Standard & \textBF{0.555} & 0.755 & \textBF{0.793} \\
\ \ \ Quantized & 0.531 & \textBF{0.760} & 0.785 \\
\bottomrule
\end{tabular}
\label{table:results}
\end{table}

Next, we assessed the performance of each approach by focusing exclusively on proteins with more than 1,024 amino acids in the test set. Table~\ref{table:results-1024} presents the results, indicating that the long and/or quantized embeddings of ESM2 T6, T12, T30, and T33 architectures achieved the highest $F_{\max}$ scores for BPO, CCO, and MFO compared to the standard configuration. For T36 embeddings, the quantized version yielded the best results for CCO and MFO. These findings lead us to conclude that ESM2 long and/or quantized embeddings are better suited for handling sequences with more than 1,024 amino acids compared to the corresponding standard models in most cases.

\begin{table}[!t]
	\centering
	\caption{$F_{\max}$ of ESM2 standard, long and quantized embeddings for proteins with more than 1,024 amino acids on the test set.}
	\begin{tabular}{lp{1.5cm}p{1.5cm}p{1.5cm}}
		\toprule
		\textBF{Method} & \textBF{BPO} & \textBF{CCO} & \textBF{MFO} \\
		\midrule
		\textbf{ESM2 T6} & & & \\
		\ \ \ Standard & 0.501 & 0.686 & 0.731 \\
		\ \ \ Long & \textBF{0.516} & \textBF{0.712} & \textBF{0.750} \\
		\ \ \ Quantized & 0.505 & 0.702 & 0.732\vspace{0.15cm} \\
		
		\textbf{ESM2 T12} & & & \\
		\ \ \ Standard & 0.493 & 0.684 & 0.743 \\
		\ \ \ Long & \textBF{0.533} & \textBF{0.698} & 0.767 \\
		\ \ \ Quantized & 0.516 & 0.693 & \textBF{0.771}\vspace{0.15cm} \\
		
		\textbf{ESM2 T30} & & & \\
		\ \ \ Standard & \textBF{0.525} & 0.698 & 0.754 \\
		\ \ \ Long & \textBF{0.525} & 0.711 & 0.758 \\
		\ \ \ Quantized & 0.506 & \textBF{0.717} & \textBF{0.771}\vspace{0.15cm} \\
		
		\textbf{ESM2 T33} & & & \\
		\ \ \ Standard & 0.517 & 0.690 & 0.761 \\
		\ \ \ Long & 0.511 & \textBF{0.718} & 0.767 \\
		\ \ \ Quantized & \textBF{0.556} & 0.716 & \textBF{0.771}\vspace{0.15cm} \\
		
		\textbf{ESM2 T36} & & & \\
		\ \ \ Standard & \textBF{0.533} & 0.717 & 0.770 \\
		\ \ \ Quantized & 0.528 & \textBF{0.733} & \textBF{0.771} \\
		\bottomrule
	\end{tabular}
	\label{table:results-1024}
\end{table}

\section{Conclusions}
\label{sec:conclusions}
In this study, we introduce an adaptation of ESM2 architectures for sequences encompassing up to 2,048 amino acids, effectively doubling the input size that the original ESM2 models can handle. In terms of the results in the protein function prediction task, the classifiers utilizing embeddings derived from long or quantized versions have outperformed the standard ESM2 configuration during our evaluation in most cases.

For future research, we highlight the adaptation of several architectures for long sequences, such as ProtT5~\cite{elnaggar2021prottrans}. Furthermore, we recognize the significance of analyzing ESM2 long and quantized architectures across different tasks. Additionally, since the long and quantized versions are pre-trained on protein data, we encourage the application of these architectures in fine-tuning processes for specific tasks, such as protein secondary structure and contact map prediction.

\section*{Acknowledgements}
The authors would like to thank the S\~ao Paulo Research Foundation (2017/12646-3), the National Council for Scientific and Technological Development (161015/2021-2, 304380/2018-0, 309330/2018-1), the Coordination~for the Improvement of Higher Education Personnel, Santander Bank - Brazil, LNCC/MCTI for providing HPC resources of the SDumont supercomputer, and Centro Nacional de Processamento de Alto Desempenho em São Paulo (CENAPAD-SP) for providing computational resources.

\bibliographystyle{sbc}
\bibliography{ref}

\end{document}